\documentclass[fleqn,twocolumn]{rovo-preprint}
\makeatletter
\providecommand\@setmarks{}
\makeatother

\begin{document}
\raggedbottom
\title{ROVO: Robust Rotation-Only Visual Odometry for UAVs at Low and Moderate Altitudes}

\author[1,2]{Xiaokai SONG}{}
\author[1,2]{Li ZHANG}{lizhang08@nudt.edu.cn}
\author[1,2]{Xichao TENG}{}
\author[1,2]{Erting PAN}{}
\author[1,2]{Yibin YE}{}
\author[3]{\protect\\ Kun WANG}{}
\author[1,2]{Leqi LIU}{}
\author[1,2]{Guangcheng DAI}{}

\address[1]{College of Aerospace Science and Engineering, National University of Defense Technology, Changsha 410000, China}
\address[2]{Hunan Provincial Key Laboratory of Image Measurement and Vision Navigation, \\  National University of Defense Technology, Changsha 410000, China}
\address[3]{Hunan Institute of Advanced Technology, Changsha 410000, China}

\abstract{Image noise can confound rotation and translation estimates in monocular visual navigation for unmanned aerial vehicles (UAVs) at low and moderate altitudes, where even small perturbations may cause large errors in translation direction. Inspired by angular cues in long-range visual perception, we propose rotation-only visual odometry (ROVO) for monocular aerial imagery. ROVO estimates pairwise rotations from matched images, robustly averages them over overlapping windows, and aligns consecutive windows through shared frames. By retaining only orientation in the trajectory state, it prevents uncertain landmark depth from propagating to later estimates. In trajectory simulation, sliding-window ROVO achieved a mean orientation error just 10.7\% as large as that of the conventional SLAM baseline. On real flight data, ROVO achieved rotation accuracy comparable to ORB-SLAM3 and DPVO under the original Bell412 settings. With a narrower FoV or at higher flight altitude, ROVO reduced relative rotation RMSE by 19.7--69.4\% compared with the next-best method in each comparison. Code and simulation tools are available at \url{https://github.com/sssxxxkkkk/ROVO}.}

\keywords{unmanned aerial vehicle, visual odometry, rotation averaging, attitude estimation, visual navigation}

\maketitle
\begingroup
\makeatletter
\long\def\@makefntext#1{\noindent#1}
\makeatother
\footnotetext{*~Corresponding author (email:~\href{mailto:lizhang08@nudt.edu.cn}{lizhang08@nudt.edu.cn})}
\endgroup

\section{Introduction}
Unmanned aerial vehicles (UAVs) rely on estimates of position and attitude for applications such as aerial surveying, mapping, delivery, and emergency response~\cite{Yan2026,2026NatSR,10603397}. A common navigation system combines measurements from a global navigation satellite system and an inertial measurement unit~\cite{drones10020091}. In urban canyons, mountainous terrain, and dense forests, satellite signals can be obstructed. During prolonged outages, errors in the inertial navigation solution accumulate, motivating the use of onboard cameras as an additional source of motion information~\cite{SCHMIDT20151}.

Visual navigation can provide either an absolute reference or a relative motion estimate~\cite{101016,Liu2025}. Scene matching navigation uses a georeferenced map to localize an onboard image~\cite{20230813626112,ye2025exploring}. Its accuracy depends on map availability and on the ability to match images acquired under different lighting conditions, in different seasons, or from different viewpoints~\cite{rs15164084}.

Relative visual navigation (RVN) methods, including visual odometry and simultaneous localization and mapping (SLAM)~\cite{174711}, estimate camera motion across images without requiring a prior georeferenced map~\cite{ABASPURKAZEROUNI2022117734}. Visual navigation has been studied for indoor point-goal navigation~\cite{Shen2025} and for low-altitude aerial visual-inertial navigation~\cite{10745211}. Extending relative visual navigation to higher altitudes raises two distinct geometric issues. For a fixed distance between camera positions, more distant terrain produces less angular parallax. A longer focal length can preserve detail on the ground but narrows the camera's field of view (FoV) for a given sensor size. When both conditions occur, a translation estimate can be sensitive to small errors in feature coordinates~\cite{1288525}. Errors in translation and depth can then propagate through triangulation and subsequent camera localization~\cite{ORBSLAM3_TRO}.

\begin{figure*}[!tb]  
    \centering 
    \includegraphics[width=\textwidth]{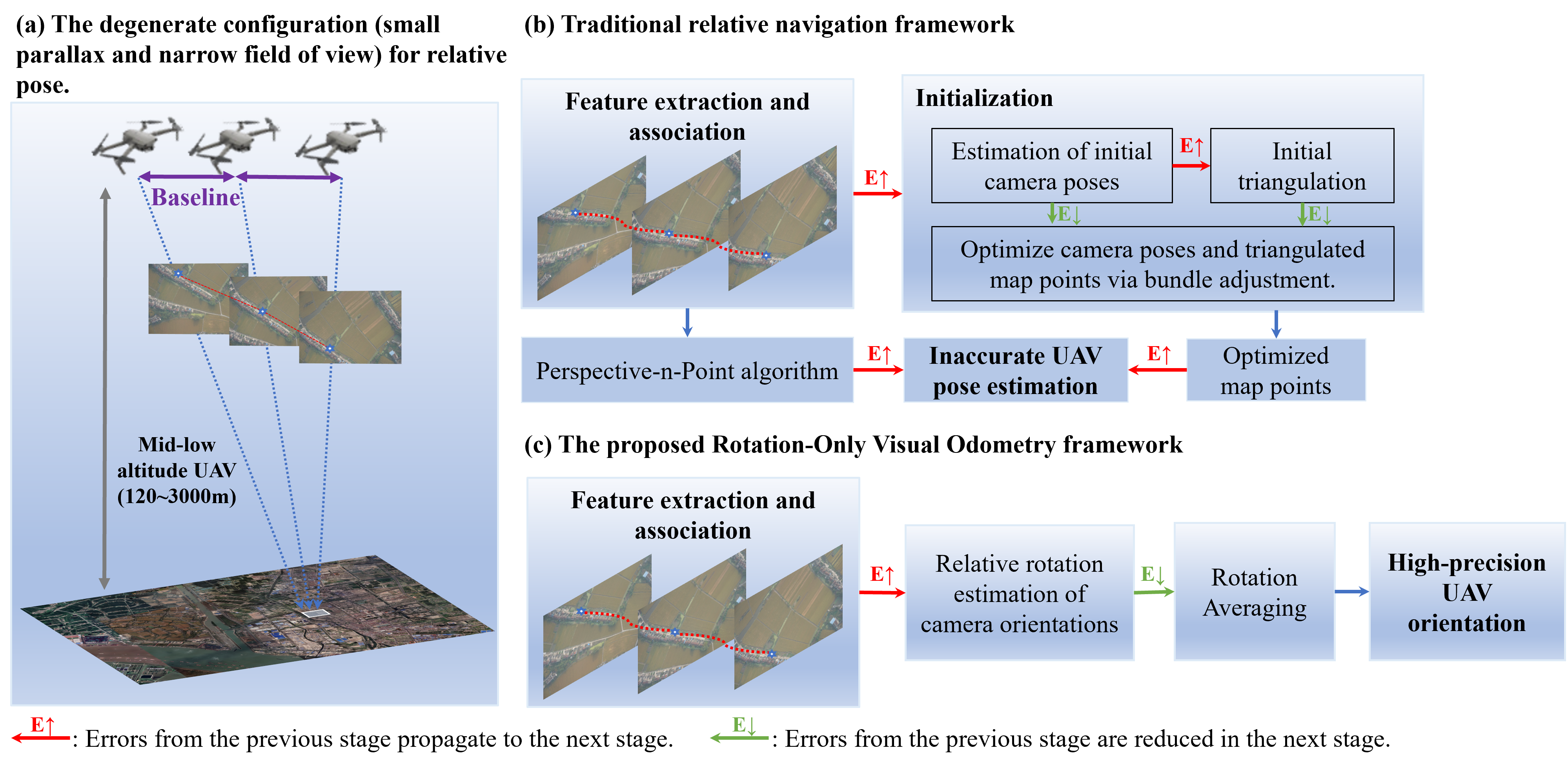}
    \caption{Relative pose estimation in aerial imagery.
    (a) A short baseline relative to scene depth gives little parallax, and a narrow FoV limits the spread of viewing directions.
    (b) In a mapping pipeline, errors in the initial pose affect triangulated landmarks and subsequent perspective-$n$-point (PnP) pose estimates.
    (c) ROVO estimates a rotation for each accepted image pair and uses rotation averaging to recover a sequence of camera orientations.}
    \label{fig:fig1} 
\end{figure*}

Rotation can remain measurable in distant scenes because a change in viewing direction still displaces image features. This observation also connects with work on the perception of self-motion from visual flow~\cite{LAPPE1999329}. We therefore propose rotation-only visual odometry (ROVO) for UAV flights at low and moderate altitudes. Figure~\ref{fig:fig1}(a) illustrates the 120--3000~m altitude range that motivates this work. Panels (b) and (c) compare a conventional mapping pipeline with ROVO. For each new frame, overlapping views provide independent predictions of its orientation. ROVO uses their agreement to choose between locally verified two-view estimates. Only the selected relative rotations enter the sliding window, which avoids propagating uncertain landmark depths or initializing a global translation scale.

This work makes three contributions.
\begin{itemize}
    \item We propose ROVO, a sliding-window framework for monocular orientation estimation. It uses robust rotation averaging within each window, aligns overlapping windows by consensus on shared frames, and fuses their estimates with bounded angular corrections.
    \item We propose a rotation-only estimator that selects among two-view rotation candidates by measuring how closely overlapping views predict the orientation of the incoming frame. The selection adapts to the available image geometry using only image measurements and previously estimated rotations.
    \item We evaluate how viewing geometry and image noise affect rotation and translation-direction estimates in simulation. Experiments on MUN-FRL and VPAIR validate ROVO under a narrow FoV and at higher flight altitudes, respectively, demonstrating improved rotation accuracy and high tracking coverage.
\end{itemize}

Section~\ref{sec:related} reviews prior work, and Section~\ref{sec:method} describes ROVO. Section~\ref{sec:sim} examines the effects of imaging geometry and noise in simulation. Section~\ref{sec:flight} evaluates the flight datasets, and Section~\ref{sec:conclusion} discusses the scope of the results.

\section{Related Work}\label{sec:related}

\subsection{Relative Visual Navigation}
Relative visual navigation estimates camera motion by associating observations across overlapping views, without requiring a georeferenced map of the environment~\cite{ABASPURKAZEROUNI2022117734}. Direct and semi-direct methods use image intensities~\cite{6906584,7219438}, whereas feature-based systems detect and match distinctive image points~\cite{ORBSLAM3_TRO,DeTone2017SuperPointSI,10377620}. Visual simultaneous localization and mapping systems jointly estimate camera poses and a map~\cite{Liu2022,10657581}, while visual-inertial systems combine visual observations with inertial measurements to support motion estimation through weak visual intervals~\cite{ORBSLAM3_TRO,8421746,4209642}. Learned odometry methods provide another way to estimate motion from image sequences~\cite{NEURIPS2023_7ac484b0}. These methods have advanced visual navigation across a range of platforms, but their estimates remain dependent on the information available in the observed scene geometry.

Aerial flight changes that geometry with altitude, baseline, and camera field of view. Liu and Shen examined initialization and sensor fusion for high-altitude monocular visual-inertial estimation~\cite{7989528}. Warren and Upcroft studied stereo visual odometry over extended-altitude flights, including the influence of camera baseline and scene depth~\cite{Warren2013HighAS,Warren2016LongrangeSV}. Later work has addressed depth uncertainty in aerial visual-inertial odometry~\cite{10.1007/978-981-96-2236-8_40}, while low-altitude flight datasets support evaluation of visual-inertial navigation in closer-range conditions~\cite{10745211}. Together, these studies show why performance should be considered in relation to flight geometry. In this work, simulation varies altitude and focal length independently, and flight experiments examine rotation accuracy under the original and narrower-field-of-view image conditions.

\subsection{Small-Parallax Geometry and Rotation Estimation}
A conventional monocular mapping pipeline estimates relative pose from image correspondences, triangulates 3D landmarks, and refines poses and landmarks through bundle adjustment~\cite{ORBSLAM3_TRO}. When the baseline is small relative to scene depth, translation induces little image motion. A narrow field of view also limits the spread of viewing directions. In these conditions, small image-coordinate errors can cause large changes in estimated translation direction and landmark depth, which can then affect later pose estimates~\cite{Warren2016LongrangeSV,HartleyZisserman2004,1288525}.

Two-view geometry has been studied through a range of relative-pose solvers, including the eight-point, seven-point, and five-point methods~\cite{601246,47116,1288525}. Direct frame-to-frame rotation optimization provides another way to estimate rotation from matched bearings~\cite{6751403}. Because solver behavior depends on image noise and viewing geometry, a low error on individual pairs does not by itself ensure consistent orientations over a long trajectory. This motivates evaluating both pairwise rotation estimates and their accumulation across frames.

Rotation-only methods reduce reliance on triangulated scene structure when estimating orientation~\cite{8793749}. Rotation averaging recovers camera orientations in a common frame by enforcing consistency among pairwise rotations, and robust formulations limit the effect of inconsistent measurements~\cite{7898474}. A hierarchical method uses cycle consistency to initialize robust rotation averaging~\cite{Lee2022HARA}. Rotation-only bundle adjustment is a related formulation that optimizes orientations from image-bearing residuals after eliminating the pairwise translation direction~\cite{9577526}. ROVO builds on this line of work with an online sliding window. It locally verifies alternative rotation estimates, selects edge measurements according to agreement among overlapping views, and combines the selected rotations through robust averaging. Translation may support local candidate checks, but only orientation is propagated in the trajectory state.

\section{Methodology}\label{sec:method}
\subsection{System Overview}\label{sec:overview}

\begin{figure*}[!tb]
    \centering
    \includegraphics[width=0.8\textwidth]{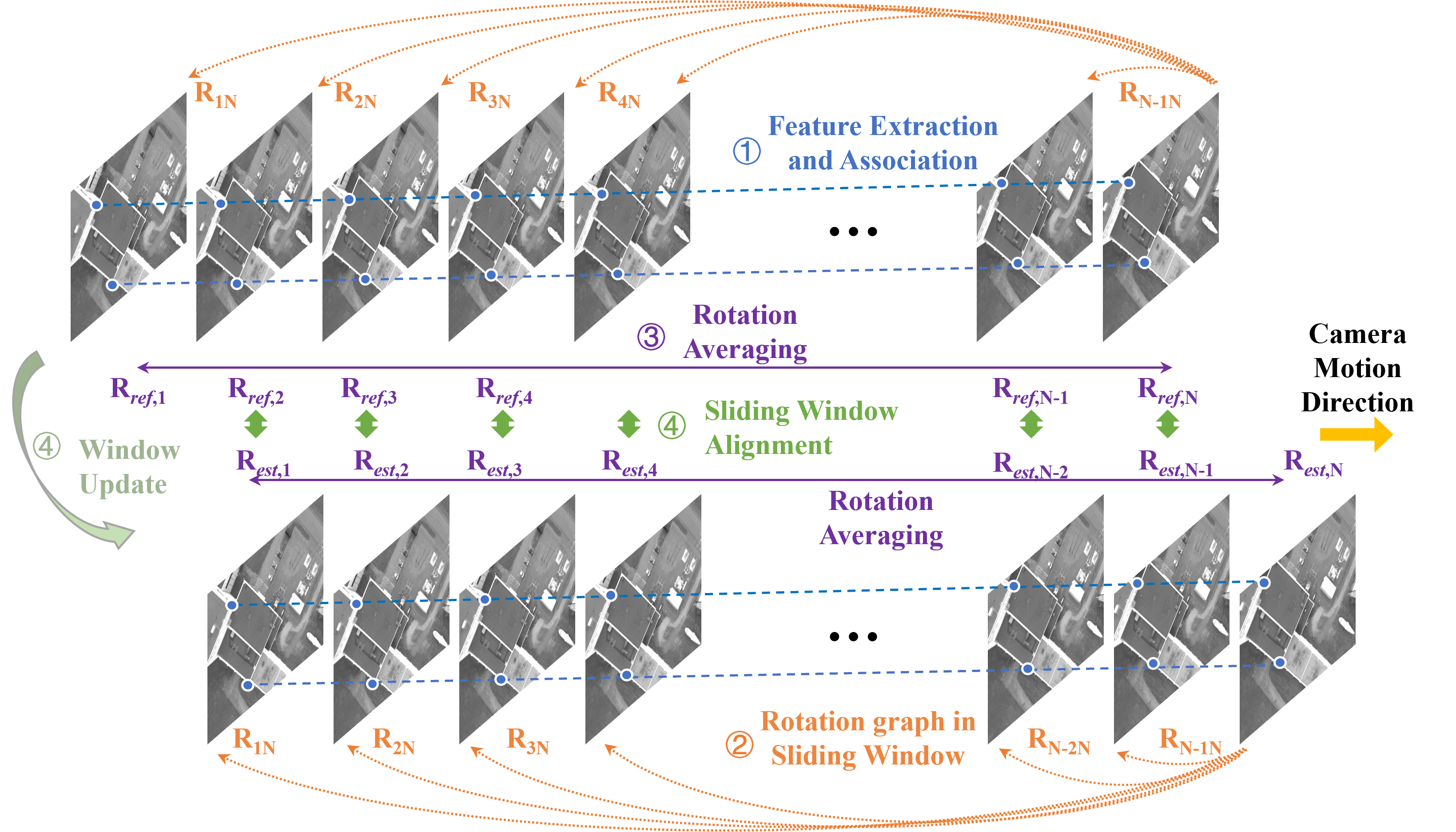}
    \caption{ROVO framework. Each orange edge represents a selected relative rotation. The blue feature matches generate the candidates, and the purple orientations are estimated by rotation averaging and aligned across windows.}
    \label{fig:fig2}
\end{figure*}
ROVO maps a calibrated monocular image stream to timestamped camera orientations (Figure~\ref{fig:fig2}). Each incoming frame is matched against the buffered keyframes to generate relative-rotation candidates. Local geometric verification tests each candidate, and joint selection uses agreement across historical views to choose the rotation measurements. After checking the selected edges for consistency, ROVO decides whether to retain the incoming frame as a keyframe. Rotation averaging then estimates the orientations within a window. Shared frames align the window to the existing trajectory, and robust fusion updates the stored orientations for subsequent frames.

The buffer contains the active keyframes and their image features. When it reaches the configured length, ROVO solves the associated rotation graph, aligns and fuses the solution, and advances the window by removing the configured number of oldest keyframes and their incident edges. The final, possibly shorter window is solved at the end of the sequence. A frame with accepted rotation edges can belong to the graph and receive an orientation even when it is not retained in the keyframe buffer.

We write $\mathbf{R}_{cw,i}$ for a rotation from world coordinates to camera $i$ and $\mathbf{R}_{wc,i}=\mathbf{R}_{cw,i}^{\top}$ for its inverse. Each tracking segment has its own world reference. A relative rotation $\mathbf{R}_{ij}$ maps camera $j$ to camera $i$, and a hat denotes an estimated relative rotation. The superscripts $\mathrm{loc}$ and $\mathrm{prev}$ distinguish estimates in the current window reference from stored estimates obtained by fusing previously processed windows. These stored historical orientations remain fixed during candidate selection for the incoming frame.

All rotation matrices belong to $\mathrm{SO}(3)$. The map $\operatorname{Log}$ returns the principal rotation vector in radians, and $\operatorname{Exp}$ maps a rotation vector to its rotation matrix. We use the angular distance
\begin{equation}
 d(\mathbf{A},\mathbf{B})=\frac{180}{\pi}
 \left\|\operatorname{Log}(\mathbf{A}^{\top}\mathbf{B})\right\|_2
\end{equation}
to express candidate disagreements in degrees. Equations involving rotation vectors use radians, with degree-valued settings converted accordingly.

\subsection{Feature Extraction and Matching}
For each incoming frame, scale-invariant feature transform (SIFT) keypoints~\cite{Burger2016} are detected in its grayscale image. If detections exceed the configured feature budget, we retain a spatially uniform subset to preserve image coverage.

Descriptor matching associates these features with each buffered keyframe. We retain a maximum one-to-one subset in which each image location is used once. An image pair with at least 15 independent matches proceeds to relative-rotation estimation. The resulting correspondences provide both the fitting data and the held-out measurements used to verify rotation candidates.

\subsection{Relative Rotation Candidates}\label{sec:two-view}
For a scene point observed in frames $i$ and $j$, let $\tilde{\mathbf{u}}_i$ and $\tilde{\mathbf{u}}_j$ be its homogeneous pixel coordinates and $\mathbf{K}$ the camera intrinsic matrix. The corresponding unit bearings are obtained by normalizing $\mathbf{K}^{-1}\tilde{\mathbf{u}}_i$ and $\mathbf{K}^{-1}\tilde{\mathbf{u}}_j$. Denoting them by $\mathbf{b}_i$ and $\mathbf{b}_j$, the epipolar constraint is~\cite{HartleyZisserman2004}
\begin{equation}
\mathbf{b}_i^{\top}[\mathbf{t}_{ij}]_{\times}
\mathbf{R}_{ij}\mathbf{b}_j=0,
\end{equation}
where $\mathbf{t}_{ij}$ is the translation direction from camera $i$ to camera $j$, expressed in camera $i$, and $[\cdot]_{\times}$ denotes the cross-product matrix. Thus the essential matrix is $\mathbf{E}_{ij}=[\mathbf{t}_{ij}]_{\times}\mathbf{R}_{ij}$. Translation directions support local verification, while the trajectory state carries only orientation.

For a pair with $n$ independent matches, we reserve a spatially distributed validation set of $m=\max(8,\lfloor n/5\rfloor)$ matches when $n\geq23$. Otherwise $m=0$. RANSAC fits the essential matrix to the remaining matches and requires at least eight fitting inliers. For the incoming frame $k$ and historical keyframe $i$, Nist\'er's five-point solver~\cite{1288525} in OpenGV~\cite{6906582} supplies a baseline rotation $\hat{\mathbf{R}}_{ik}^{(0)}$. Essential-matrix decompositions from the RANSAC initializer and a normalized linear estimate supply additional hypotheses. We rank these hypotheses by support and break ties by median reprojection error. The chosen baseline initializes the iterative eigensolver of Kneip and Lynen~\cite{6751403}, which produces a refined candidate $\hat{\mathbf{R}}_{ik}^{(1)}$ from the same fitting bearings.

Both candidates are tested on the same fitting and validation sets. With weak parallax, a match supports a candidate when its derotated bearings agree within the angular tolerance corresponding to two pixels. When translation directions estimated from two spatial subsets agree within $15^\circ$ up to sign, we also test positive depth and two-view reprojection error with a two-pixel threshold. The fitting data determine the translation sign. A locally valid candidate must be supported by at least eight fitting inliers and at least half of the fitting inliers. When $m>0$, it must also have at least $\min(8,\lceil m/2\rceil)$ supporting validation matches. Its support count is the total number of supporting fitting and validation matches.

\subsection{Joint Selection of Rotation Measurements}\label{sec:selection}
Local image support can leave both candidates plausible. We resolve this ambiguity by requiring rotations estimated against different historical keyframes to predict a consistent orientation for the same incoming frame. A candidate is evaluated together with the candidates on the other incoming edges, so that a consistent group of refinements can be selected jointly.

For a historical frame $i$ with a stored orientation, candidate $a\in\{0,1\}$ predicts the current camera-to-world orientation
\begin{equation}
\mathbf{Q}_{k\mid i}^{(a)}=
\mathbf{R}_{wc,i}^{\mathrm{prev}}\hat{\mathbf{R}}_{ik}^{(a)}.
\end{equation}
For historical frames $i$ and $j$ in the same tracking segment, define $\mathbf{H}_{ij}=(\mathbf{R}_{wc,i}^{\mathrm{prev}})^{\top}\mathbf{R}_{wc,j}^{\mathrm{prev}}$. This rotation maps camera $j$ to camera $i$. The disagreement between candidates $a$ and $b$ is
\begin{equation}
\epsilon_{ij}^{(k)}(a,b)=d\bigl(
\hat{\mathbf{R}}_{ik}^{(a)},
\mathbf{H}_{ij}\hat{\mathbf{R}}_{jk}^{(b)}\bigr).
\end{equation}
It equals $d(\mathbf{Q}_{k\mid i}^{(a)},\mathbf{Q}_{k\mid j}^{(b)})$ when both stored orientations are available. During startup or before a historical frame has been fused, an accepted historical edge can instead supply $\mathbf{H}_{ij}$. This closes a three-frame loop and permits selection before the first window solution. Pairs from different known segments, or without either historical relation, are omitted.

Let $\mathcal{P}_k$ contain the comparable pairs of historical keyframes whose incoming edges both have locally valid baselines. A binary label $s_i$ selects candidate 0 or 1 on edge $(i,k)$, and $\mathbf{s}$ collects these labels. We minimize
\begin{equation}
J_k(\mathbf{s})=\sum_{(i,j)\in\mathcal{P}_k}
\rho_{\delta_{\mathrm{sel}}}\bigl(
\epsilon_{ij}^{(k)}(s_i,s_j)\bigr),
\end{equation}
where $\rho_{\delta}$ is the Huber loss, equal to $x^2/2$ for a nonnegative residual $x\leq\delta$ and to $\delta(x-\delta/2)$ otherwise. The selection scale is $\delta_{\mathrm{sel}}=3^\circ$, and the residuals and scale are evaluated in degrees. Each edge has one label shared across all terms, which enforces consistency among its comparisons.

The joint selector considers a refinement only when its edge has at least eight held-out matches. The search enumerates the eligible binary labels. An edge with two locally valid candidates can switch to the refinement when it has at least two other comparable edges with valid baselines. For each eligible edge, we minimize $J_k$ with its label fixed first to 0 and then to 1, optimizing the other labels in each case. A refinement remains eligible only if this cost reduction, divided by the number of comparable baseline-supported neighbors, exceeds $0.1~\mathrm{deg}^2$. We fix ineligible edges with valid baselines to label 0 and minimize over the remaining eligible labels. An isolated edge retains a locally valid baseline. When only the refinement passes local verification, it is accepted if it agrees within $8^\circ$ with at least two selected candidates on baseline-supported neighboring edges. These neighbors must also survive the consistency check described next.

\subsection{Sliding-Window Rotation Averaging}\label{sec:rotation-graph}
\subsubsection{Graph Update and Keyframe Selection}
After joint selection, we check a nonconsecutive edge against the product of the intervening consecutive rotations. A complete chain rejects the direct edge when every link has at least as much support as that edge and their angular disagreement exceeds $8^\circ$. The surviving edges and their support counts determine the graph update and keyframe decision.

The first frame initializes the keyframe buffer. An incoming frame becomes a keyframe if it has no surviving rotation edge or if the largest support count on its surviving edges falls below the keyframe threshold. An unmatched keyframe is retained so that later frames can connect it to the graph. A frame with sufficient support contributes its accepted constraints as a graph vertex outside the keyframe buffer. The window is solved and advanced according to the schedule in Section~\ref{sec:overview}.

\subsubsection{Rotation-Averaging Objective}
Let $\mathcal{G}=(\mathcal{V},\mathcal{E})$ denote the connected component containing the highest-degree vertex of the updated graph. Its vertices are frame indices, and its edges carry the selected measurements $\hat{\mathbf{R}}_{ij}$. Within the current window reference, these measurements satisfy
\begin{equation}
\hat{\mathbf{R}}_{ij}\approx
\mathbf{R}_{cw,i}^{\mathrm{loc}}
\bigl(\mathbf{R}_{cw,j}^{\mathrm{loc}}\bigr)^{\top}.
\end{equation}
For each edge, define the rotation-vector residual
\begin{equation}
\mathbf{r}_{ij}=\operatorname{Log}\!\left(
\bigl(\mathbf{R}_{cw,i}^{\mathrm{loc}}\bigr)^{\top}
\hat{\mathbf{R}}_{ij}\mathbf{R}_{cw,j}^{\mathrm{loc}}
\right).
\end{equation}
Let $\mathcal{E}_{\mathrm{in}}\subseteq\mathcal{E}$ be the edges retained after the initialization and filtering below. We fix the first retained frame $r$ at its initialized rotation $\bar{\mathbf{R}}_{cw,r}^{\mathrm{loc}}$ and solve
\begin{equation}
\begin{aligned}
\underset{\{\mathbf{R}_{cw,i}^{\mathrm{loc}}\}_{i\in\mathcal{V}\setminus\{r\}}}
{\operatorname{argmin}}\quad &
\sum_{(i,j)\in\mathcal{E}_{\mathrm{in}}}
\|\mathbf{r}_{ij}\|_2^{1/2},\\
\text{subject to}\quad &
\mathbf{R}_{cw,r}^{\mathrm{loc}}=
\bar{\mathbf{R}}_{cw,r}^{\mathrm{loc}}.
\end{aligned}
\end{equation}
Fixing the anchor removes the common rotation ambiguity. The $\ell_{1/2}$ loss limits the influence of inconsistent edges, and iteratively reweighted least squares (IRLS) refines the initialized orientations.

\subsubsection{Initialization and Robust Refinement}
We use the hierarchical approach for robust rotation averaging (HARA)~\cite{Lee2022HARA} to initialize each connected window. It samples up to ten common neighbors per edge. For an edge $(i,j)$ and a sampled common neighbor $\ell$, the dimensionless three-view consistency error is
\begin{equation}
\zeta_{ij\ell}=\left\|\hat{\mathbf{R}}_{ij}-
\hat{\mathbf{R}}_{i\ell}
\hat{\mathbf{R}}_{j\ell}^{\top}\right\|_F^2.
\end{equation}
HARA derives its initialization thresholds from the 10th, 20th, and 30th percentiles of sampled errors below 1. Starting from the highest-degree vertex at the identity rotation, it propagates orientations through edges supported by consistent loops. The initial loop-support requirement of ten is relaxed when propagation stalls. For any remaining vertices, initialized neighbors vote for the next vertex to add. The vertex with the most votes receives the chordal $L_1$ mean of their predicted rotations. If fewer than five sampled loops pass the initial test, edge propagation supplies the initial orientations and all edges proceed to refinement.

For a HARA-initialized window, we filter edges only when the median of all sampled loop errors is at most 1. We then retain an edge if the squared Frobenius difference between its measurement and the initialized endpoint prediction is below 1. Otherwise all edges are retained. These dimensionless checks determine $\mathcal{E}_{\mathrm{in}}$. IRLS uses quaternions for at most 250 iterations and stops when the mean angular update falls below $10^{-3}$ radians.

After refinement, each frame $i$ receives a dimensionless quality score $q_i$ based on its total incident match support, graph degree, and median angular residual. These quantities are evaluated on $\mathcal{E}$, including edges removed by HARA filtering. Scores are normalized by their positive median within the window and clipped to $[0.05,4]$. The resulting $q_i$ favors orientations constrained by multiple well-supported, consistent measurements and supplies the weights for inter-window alignment and fusion.

\subsection{Inter-Window Alignment and Orientation Fusion}\label{sec:fusion}
\subsubsection{Alignment from Shared Frames}
Rotation averaging returns orientations in the current window reference. Shared frames relate this reference to a previously estimated trajectory segment. We first choose the segment with the largest total alignment weight among the shared frames. Let $\mathcal{O}$ contain its shared frame indices. For each $i\in\mathcal{O}$, $\mathbf{R}_{cw,i}^{\mathrm{loc}}$ is the local estimate and $\mathbf{R}_{wc,i}^{\mathrm{prev}}$ is its stored camera-to-world orientation. Their candidate alignment rotation is
\begin{equation}
\mathbf{G}_i=
\bigl(\mathbf{R}_{cw,i}^{\mathrm{loc}}\bigr)^{\top}
\bigl(\mathbf{R}_{wc,i}^{\mathrm{prev}}\bigr)^{\top},
\end{equation}
which maps the segment world reference into the window reference. The alignment weight is $w_i=q_i\max(I_i^{\mathrm{prev}},I_{\min})$, where $I_i^{\mathrm{prev}}$ is the stored information weight. The positive constants $I_{\min}$ and $I_{\max}$ provide an information floor before robust weighting and an upper cap on stored information, respectively.

For a candidate alignment $\mathbf{G}_{\ell}$, with $\ell\in\mathcal{O}$, define its consensus set
\begin{equation}
\mathcal{C}_{\ell}=\left\{i\in\mathcal{O}:
\left\|\operatorname{Log}(\mathbf{G}_i\mathbf{G}_{\ell}^{\top})\right\|_2
\leq\tau_{\mathrm{align}}\right\},
\end{equation}
where $\tau_{\mathrm{align}}$ is the angular consensus threshold in radians. We choose the hypothesis with the largest $\sum_{i\in\mathcal{C}_{\ell}}w_i$, breaking ties by the weighted sum of angular errors truncated at $\tau_{\mathrm{align}}$. Let $\mathcal{C}$ denote the winning consensus set. Its weighted matrix mean, projected onto $\mathrm{SO}(3)$ by singular value decomposition, initializes the alignment $\Delta\mathbf{R}$. Up to 30 robust iterations refine it using
\begin{equation}
\begin{aligned}
\mathbf{v}_i&=\operatorname{Log}(\mathbf{G}_i\Delta\mathbf{R}^{\top}),\\
\bar w_i&=w_i h(\|\mathbf{v}_i\|_2;\sigma_{\mathrm{align}}),\\
\Delta\mathbf{R}&\leftarrow\operatorname{Exp}\!\left(
\frac{\sum_{i\in\mathcal{C}}\bar w_i\mathbf{v}_i}
{\sum_{i\in\mathcal{C}}\bar w_i}\right)\Delta\mathbf{R}.
\end{aligned}
\end{equation}
Here $\mathbf{v}_i$ is an alignment residual in radians, $\sigma_{\mathrm{align}}$ is the robust angular scale, and the Huber weight satisfies $h(x;\sigma)=1$ for $0\leq x\leq\sigma$ and $h(x;\sigma)=\sigma/x$ for $x>\sigma$. Applying the alignment to any solved frame gives
\begin{equation}
\mathbf{R}_{wc,i}^{\mathrm{cand}}=
\bigl(\mathbf{R}_{cw,i}^{\mathrm{loc}}\Delta\mathbf{R}\bigr)^{\top},
\end{equation}
where $\mathrm{cand}$ denotes an aligned candidate in the segment world reference.

\subsubsection{Bounded Orientation Fusion}
For a frame with a stored estimate in the selected segment, the candidate supplies a new observation. Its residual is $\mathbf{e}_i=\operatorname{Log}((\mathbf{R}_{wc,i}^{\mathrm{prev}})^{\top}\mathbf{R}_{wc,i}^{\mathrm{cand}})$. We assign this observation an information weight $I_i^{\mathrm{obs}}$ and compute its fusion fraction $\beta_i$ as
\begin{equation}
\begin{aligned}
I_i^{\mathrm{obs}}={}&
\min\!\bigl(\max(q_i,I_{\min}),I_{\max}\bigr)\\
&\cdot h(\|\mathbf{e}_i\|_2;\sigma_{\mathrm{fuse}}),\\
\beta_i={}&\frac{I_i^{\mathrm{obs}}}
{\max(I_i^{\mathrm{prev}},I_{\min})+I_i^{\mathrm{obs}}},
\end{aligned}
\end{equation}
where $\sigma_{\mathrm{fuse}}$ is the fusion scale in radians. A maximum correction angle $\theta_{\max}$ bounds the update through
\begin{equation}
\alpha_i=\begin{cases}
\beta_i, & \|\mathbf{e}_i\|_2=0,\\[4pt]
\min\!\left(\beta_i,\dfrac{\theta_{\max}}{\|\mathbf{e}_i\|_2}\right),
& \|\mathbf{e}_i\|_2>0,
\end{cases}
\end{equation}
with $\theta_{\max}$ also expressed in radians. The updated orientation and accumulated information are
\begin{equation}
\begin{aligned}
\mathbf{R}_{wc,i}^{\mathrm{new}}={}&
\mathbf{R}_{wc,i}^{\mathrm{prev}}
\operatorname{Exp}(\alpha_i\mathbf{e}_i),\\
I_i^{\mathrm{new}}={}&\min\!\bigl(I_{\max},
\max(I_i^{\mathrm{prev}},I_{\min})+I_i^{\mathrm{obs}}\bigr).
\end{aligned}
\end{equation}
The information cap limits the influence of correlated estimates from overlapping windows while preserving the ability to accept later corrections. The superscript $\mathrm{new}$ denotes the stored state after this update. It becomes the $\mathrm{prev}$ state when subsequent frames are processed.

Frames without a stored estimate in the selected segment are initialized from their aligned candidates, with information $\min(\max(q_i,I_{\min}),I_{\max})$. A solved window with no valid overlap starts a new trajectory segment using the identity alignment. These stored orientations supply the historical predictions in Section~\ref{sec:selection} and form the timestamped output trajectory.

\section{Simulation Experiments}\label{sec:sim}
\subsection{Simulation Setup}

\begin{figure}[!tb]
    \centering
    \includegraphics[width=\linewidth]{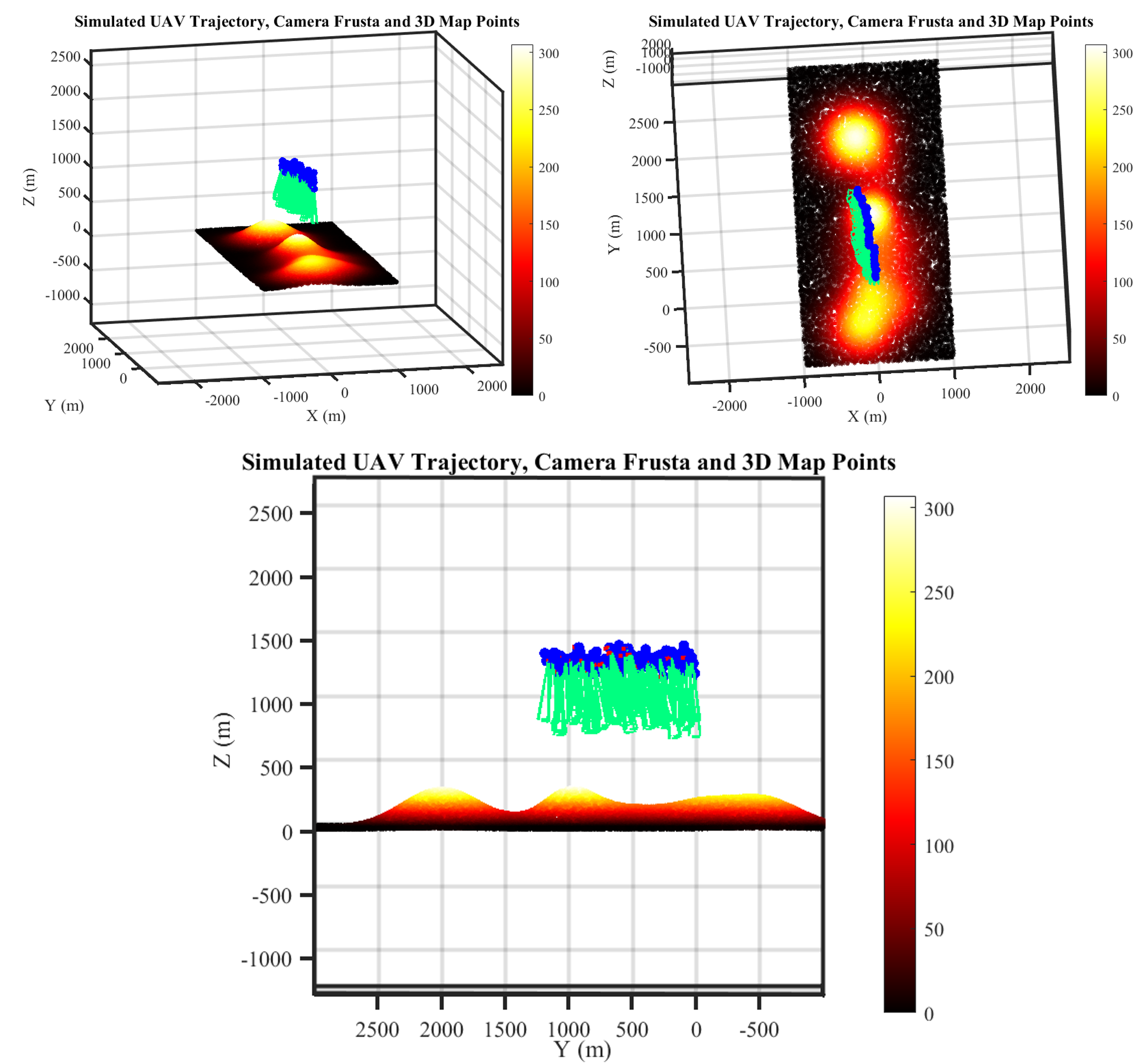}
    \caption{Simulated flight trajectory and camera viewing frusta above the terrain.}
    \label{fig:fig3}
\end{figure}
The simulated terrain is a point cloud with hills up to 300~m high and Gaussian surface-height perturbations with a standard deviation of 0.5~m. Each point has a unique identifier so that correspondences are known even when noise is added to image coordinates.

The nominal flight altitude is 1300~m, with Gaussian vertical perturbations of standard deviation 50~m. The UAV advances along the positive $Y$-axis with a mean step of 10~m and Gaussian step perturbations of standard deviation 3~m. Its lateral displacement increments have a Gaussian standard deviation of 10~m. At each step, rotations of up to $\pm3^\circ$ about the three body axes are accumulated. The simulated camera has a resolution of $640\times512$ pixels and a nominal focal length of 3600 pixels.

Let $\mathbf{X}_w=[X_w,Y_w,Z_w,1]^\top$ be a point in homogeneous world coordinates. With camera orientation $\mathbf{R}_{wc}$ and camera center $\mathbf{t}_{wc}$ expressed in the world frame, the point has camera coordinates
\begin{equation}
\mathbf{X}_c = \begin{bmatrix} \mathbf{R}_{wc}^\top & -\mathbf{R}_{wc}^\top \mathbf{t}_{wc} \end{bmatrix} \mathbf{X}_w,
\end{equation}
where $\mathbf{X}_c=[X_c,Y_c,Z_c]^\top$. For a point in front of the camera, the perspective projection is
\begin{equation}
\mathbf{u} = \mathbf{K} \frac{\mathbf{X}_c}{Z_c},
\end{equation}
where $\mathbf{u}=[u,v,1]^\top$ denotes homogeneous pixel coordinates. Figure~\ref{fig:fig3} shows how the viewing frusta intersect the terrain along the simulated flight path.

\subsection{Error Metrics}
We measure the angular difference between an estimated rotation $\mathbf{R}_{\mathrm{est}}$ and its ground truth $\mathbf{R}_{\mathrm{gt}}$ in degrees,
\begin{equation}
\theta_{\text{rot}} = \operatorname{rad2deg}\!\left( \arccos\!\left( \frac{\operatorname{trace}\left( \mathbf{R}_{\text{est}}^\top \mathbf{R}_{\text{gt}} \right) - 1}{2} \right) \right).
\end{equation}

For the estimated and reference translation directions, $\mathbf{T}_{\mathrm{est}}$ and $\mathbf{T}_{\mathrm{gt}}$, we use the smaller angle obtained for either sign of the estimate,
\begin{equation}
\phi_{\text{trans}} = \operatorname{rad2deg}\!\left( \arccos\!\left( \frac{\bigl| \mathbf{T}_{\text{est}}^\top \mathbf{T}_{\text{gt}} \bigr|}{\|\mathbf{T}_{\text{est}}\| \|\mathbf{T}_{\text{gt}}\|} \right) \right).
\end{equation}

In the two-view experiments, averaging $\theta_{\mathrm{rot}}$ and $\phi_{\mathrm{trans}}$ over image pairs gives the mean rotation error (MRE) and mean translation-direction error (MTE), respectively. 

\subsection{Comparison of Relative Rotation Solvers}\label{sec:sim-solvers}
We evaluate the two-view solvers from OpenGV~\cite{6906582} using the nominal flight and camera configuration above. Zero-mean Gaussian noise is added to the projected feature coordinates, with standard deviations from 0 to 3 pixels. The simulation supplies known feature correspondences. The solver calls use a perturbed reference pose, and the reference rotation is used to select among candidate solutions. This protocol evaluates solver sensitivity to image noise with controlled correspondences and reference-assisted candidate selection.

\begin{figure}[!tb]  
    \centering 
    \includegraphics[width=\linewidth]{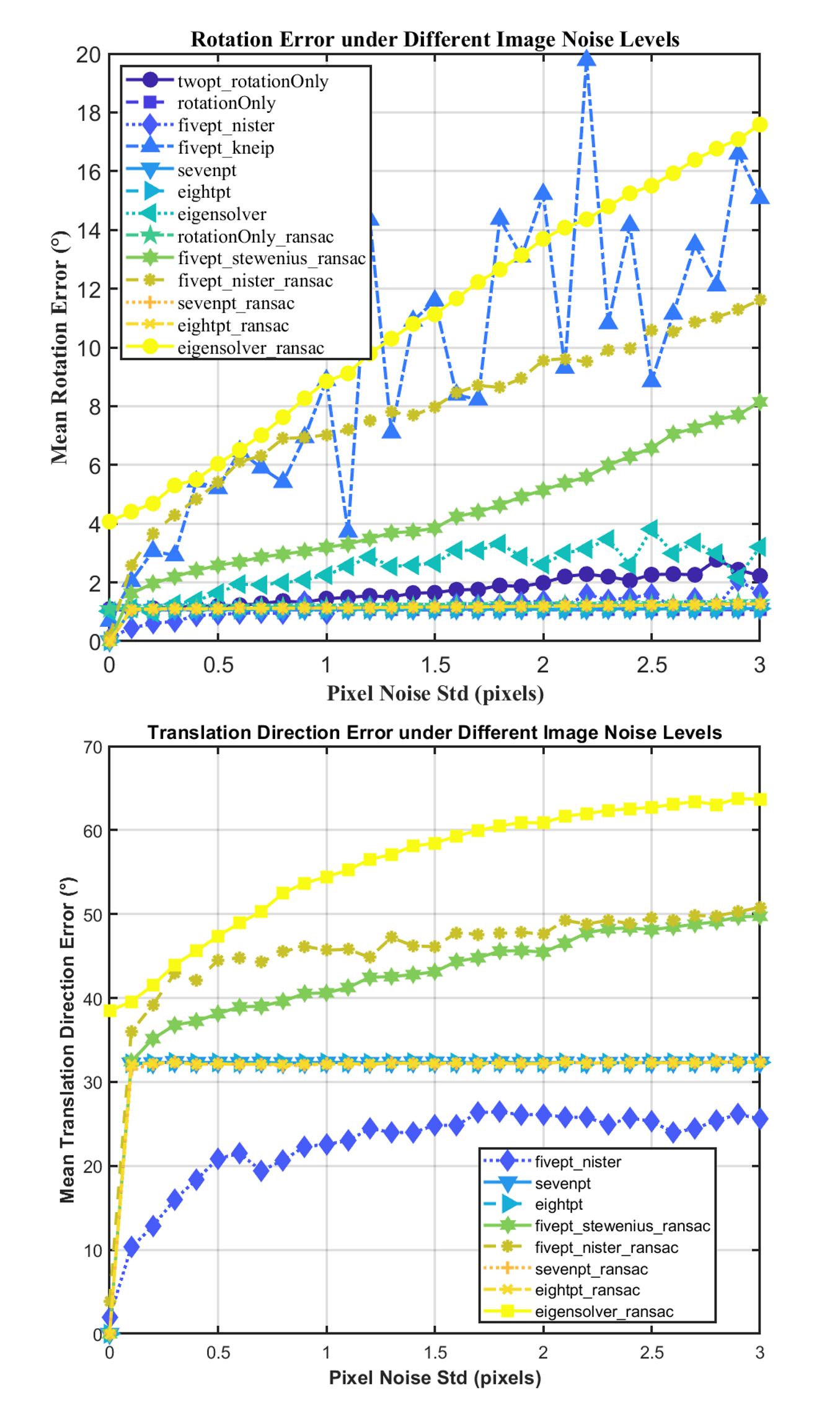}
    \caption{Two-view rotation and translation-direction errors for each solver as the standard deviation of Gaussian pixel noise increases.}
    \label{fig:fig4} 
\end{figure}

Figure~\ref{fig:fig4} shows that a small rotation error can coexist with a much larger translation-direction error for the same correspondences. At a noise standard deviation of 3 pixels, Nist\'er's five-point solver gives a mean two-view rotation error of $1.65^\circ$ and a translation-direction error of $25.61^\circ$. The eigensolver gives a two-view rotation error of $3.21^\circ$ at the same noise level.

Rotation changes the direction of each viewing ray, whereas the image displacement due to translation depends on the baseline relative to scene depth. When this ratio is small, an image perturbation can obscure translational parallax while leaving rotation measurable. The results support estimating orientation independently when translation direction is poorly constrained, while the different solver errors motivate retaining alternative rotation candidates.

\subsection{Influence of Flight Altitude and Camera Focal Length}
We vary nominal altitude and focal length separately at a fixed image-noise standard deviation of 1 pixel. The altitude sweep covers 580 to 1300~m in steps of 80~m, and the focal-length sweep covers 900 to 3600 pixels in steps of 300 pixels. The remaining configuration and the two-view evaluation protocol follow the preceding experiment.

\begin{figure}[!tb]
    \centering
    \includegraphics[width=\linewidth]{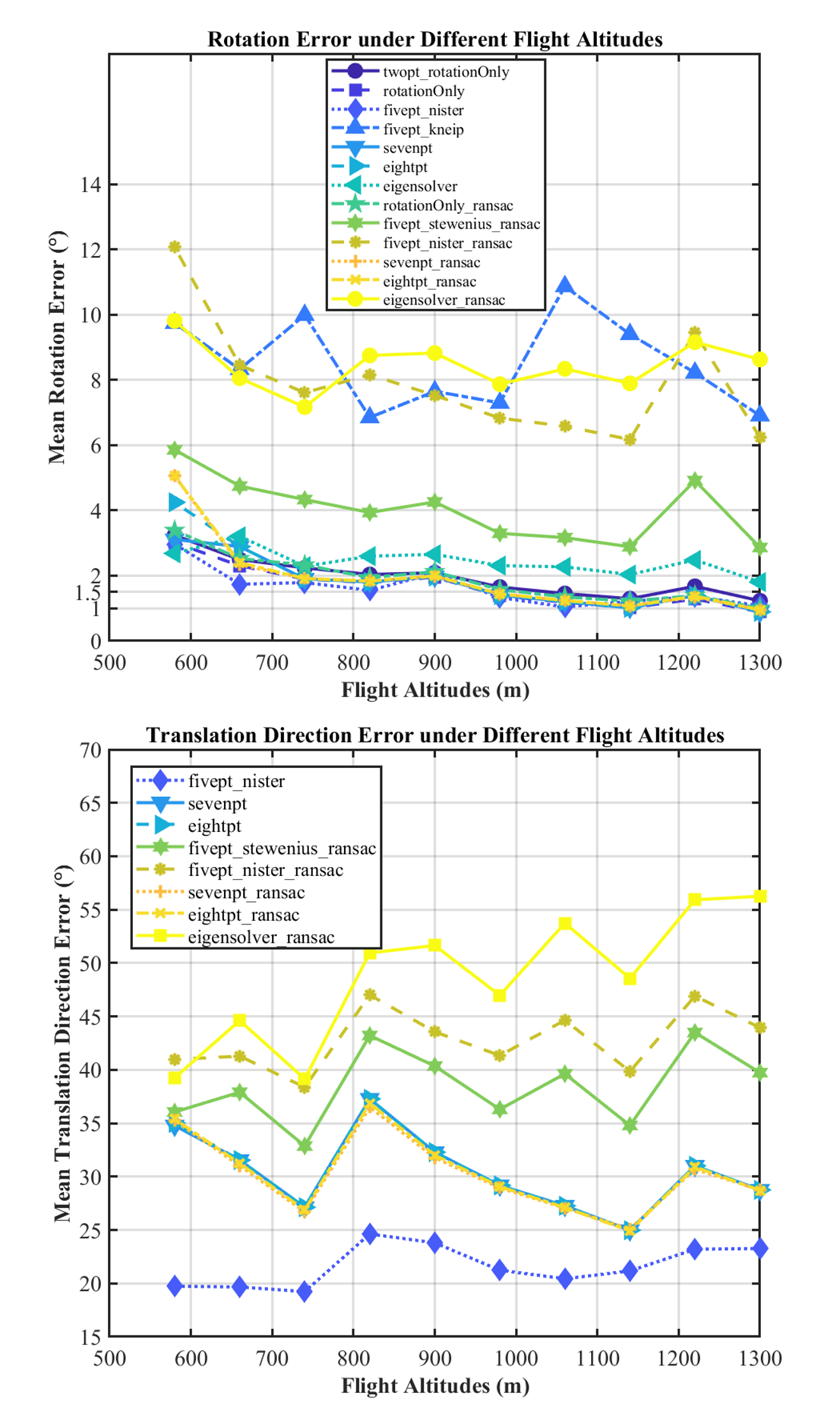}
    \caption{Two-view rotation and translation-direction errors versus simulated flight altitude at a fixed camera focal length and an image-noise standard deviation of 1 pixel.}
    \label{fig:fig5}
\end{figure}

In the altitude sweep (Figure~\ref{fig:fig5}), Nist\'er's rotation error decreases between the endpoints even as its translation-direction error increases. Other solvers show different trends across the sampled altitudes.

\begin{figure}[!tb]
    \centering
    \includegraphics[width=\linewidth]{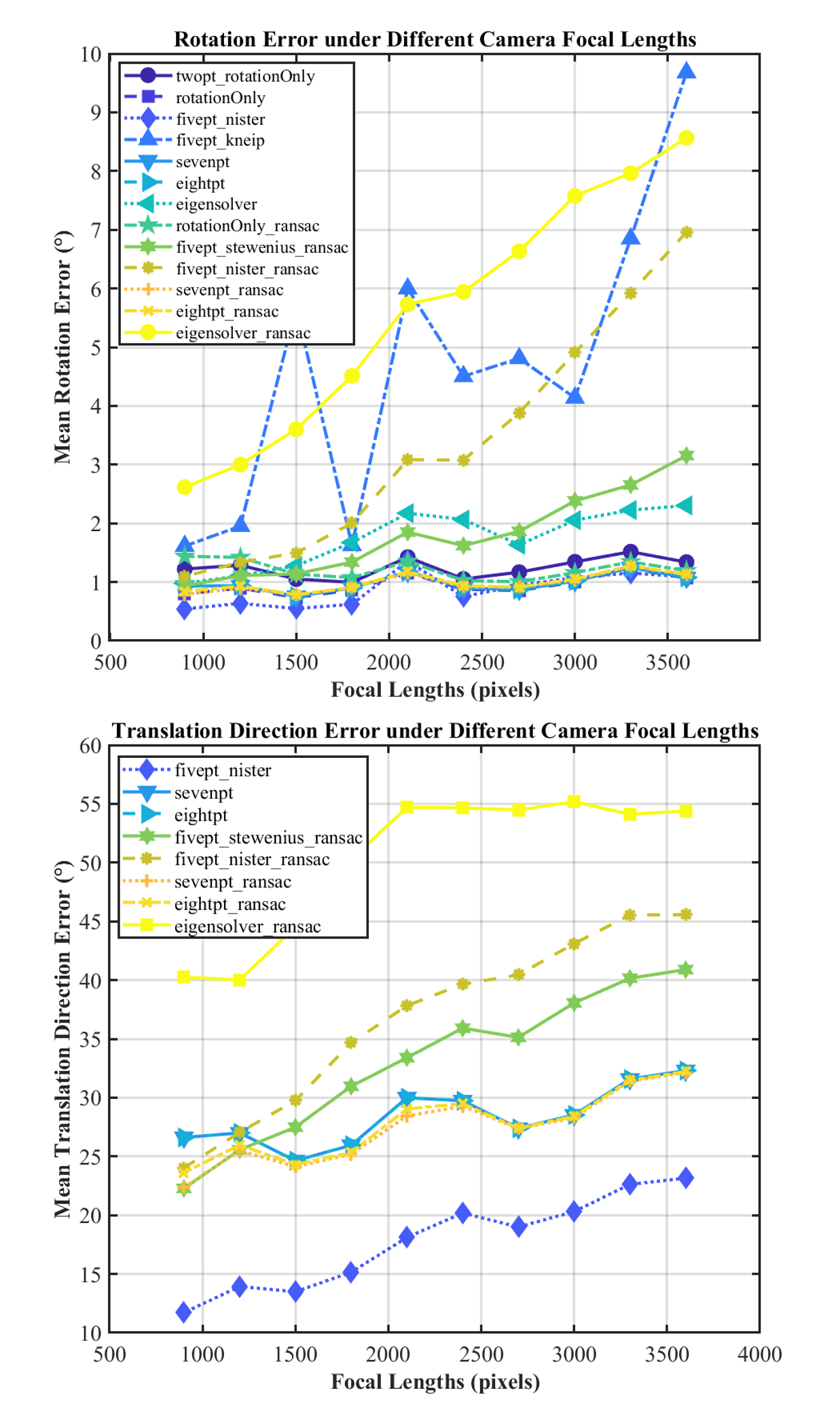}
    \caption{Two-view rotation and translation-direction errors versus camera focal length at a fixed simulated flight altitude and an image-noise standard deviation of 1 pixel.}
    \label{fig:fig6}
\end{figure}

Figure~\ref{fig:fig6} shows that translation-direction error generally increases as the FoV narrows. Nist\'er's solver improves in both MRE and MTE with a wider FoV.

Increasing altitude at a fixed baseline reduces the ratio of baseline to scene depth. The resulting loss of translational parallax makes translation direction more sensitive to image noise, while the viewing rays approach those of a pure rotation. This explains how rotation accuracy can improve as translation-direction accuracy deteriorates in the altitude sweep.

Changing focal length affects both the angular noise and the spread of viewing directions. A smaller focal length increases the angular perturbation represented by a fixed pixel error and admits bearings over a wider range of directions. In the simulated scene, this wider spread helps separate the image motions induced by rotation and translation, which is consistent with the observed improvement at the wider FoV. The differing solver responses motivate selecting rotations through agreement across views.

\subsection{Simulation Comparison with a Conventional SLAM System}
The simulated SLAM baseline initializes with structure from motion, associates known features with 3D points, and uses PnP and sliding-window BA to update the poses and landmarks. The rotation graph uses correspondences from synthetic point identifiers. The solver calls follow the reference-assisted protocol in Section~\ref{sec:sim-solvers}. Pixel-noise standard deviation is swept over the same range, with the two-view solver held fixed within each run. Joint candidate selection is evaluated separately on flight images in Section~\ref{sec:flight}.

For each rotation solver, we compare three aggregation schemes. Global rotation averaging (RA) uses all available frames in one graph. Windowed RA uses overlapping windows of at most ten keyframes and advances by one keyframe after each full-window update. Chaining composes consecutive pairwise rotations in temporal order. These configurations measure how pairwise solver errors affect trajectory orientation after initial-frame alignment.

\begin{figure}[!tb]  
    \centering 
    \includegraphics[width=\linewidth]{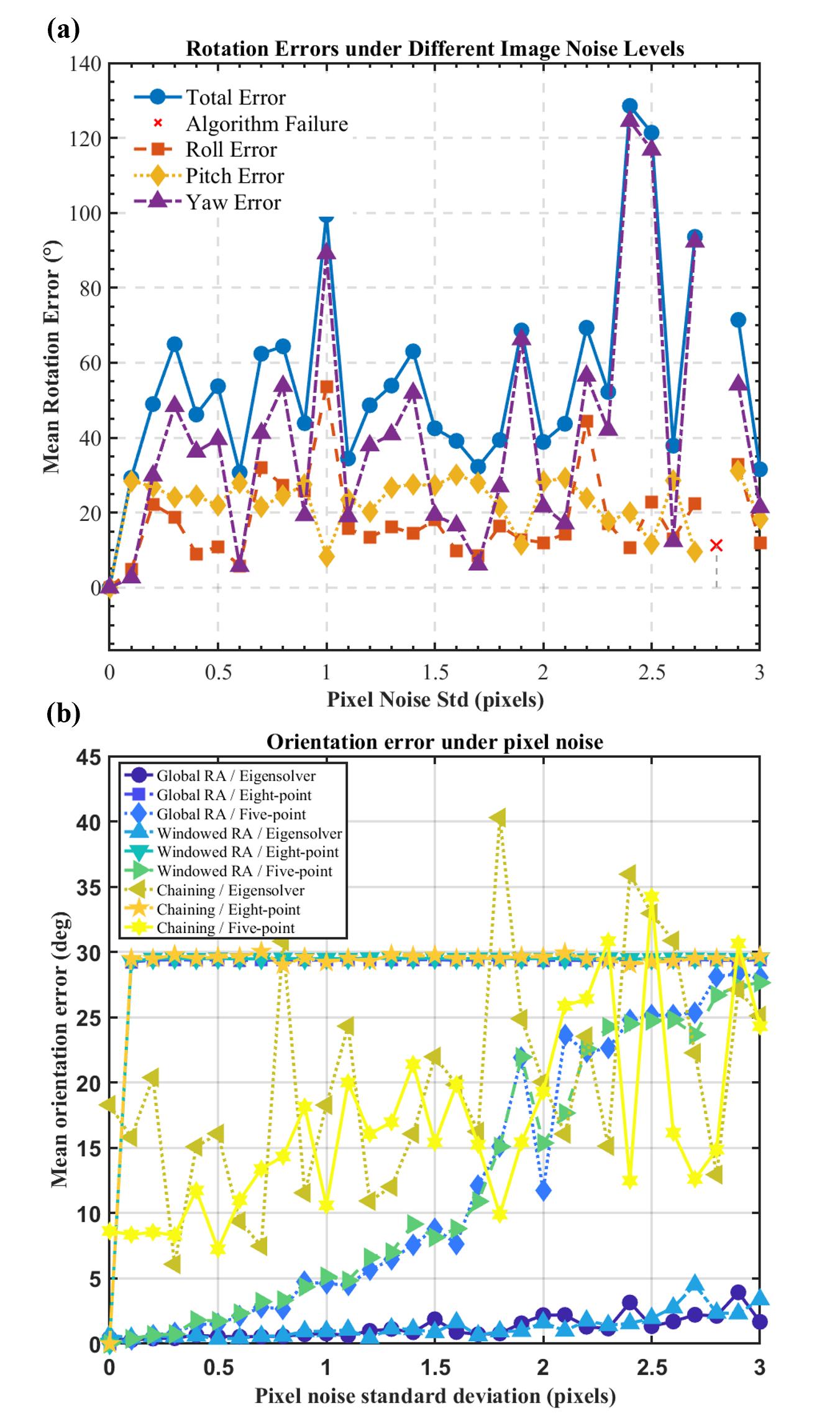}
    \caption{Trajectory orientation errors versus pixel-noise standard deviation. (a) Total and Euler-angle component errors for the simulated SLAM baseline. Crosses mark failures. (b) Three fixed two-view solvers evaluated with full-sequence rotation averaging (Global RA), sliding-window rotation averaging (Windowed RA), and direct composition of consecutive rotations (Chaining). Five-point denotes Nist\'er's solver.}
    \label{fig:fig7}
\end{figure}

The SLAM baseline in Figure~\ref{fig:fig7}(a) shows sharp fluctuations in orientation error, with yaw following several of the largest excursions in the total error. Figure~\ref{fig:fig7}(b) shows that rotation averaging substantially reduces accumulated error for the eigensolver. At a noise standard deviation of 3 pixels, windowed and global RA give orientation errors of $3.39^\circ$ and $1.68^\circ$, respectively, compared with $25.12^\circ$ for chaining. The five-point solver has the lower two-view rotation error in Figure~\ref{fig:fig4}, but its windowed trajectory error reaches $27.65^\circ$. The eight-point trajectory errors cluster near $29^\circ$ once noise is introduced.

With small triangulation angles, perturbations in the input bearings can produce large changes in reconstructed depths. PnP estimates the next orientation against that uncertain map, and the resulting pose contributes to later map updates. This coupling provides a path through which image noise can affect subsequent orientation estimates.

Chaining carries each pairwise error into later orientations. Rotation averaging constrains each orientation through overlapping paths, allowing inconsistent edges to be identified and isolated deviations to be downweighted. Smaller window graphs retain much of this benefit. The different solver rankings in the two-view and trajectory experiments show that trajectory accuracy also depends on consistency across views. This dependence motivates the multi-view rotation selection in Section~\ref{sec:selection}.

\section{Experimental Evaluation}\label{sec:flight}
\subsection{Flight Datasets and Evaluation Protocol}
We compare our MATLAB R2023b implementation of ROVO with monocular ORB-SLAM3~\cite{ORBSLAM3_TRO} and Deep Patch Visual Odometry (DPVO)~\cite{NEURIPS2023_7ac484b0}. The MUN-FRL dataset~\cite{10.1177/02783649241238358} provides visual, inertial, and lidar measurements from a Bell 412 helicopter together with a postprocessed navigation reference. Its downward-looking camera records $1440\times1080$ images at 20~Hz. We use three flights over different ground scenes. For a second imaging condition, we reproject the same views with both pixel focal lengths ($f_x$ and $f_y$) multiplied by three and update each estimator's intrinsics accordingly. This tests a narrower FoV at the recorded camera positions. The camera-to-ground distance and the baseline between exposures are the same in both conditions.

VPAIR, a dataset for aerial visual place recognition and localization~\cite{schleiss2022vpair}, provides images from a light aircraft flying over urban areas, farmland, and forests. The flight altitude reported in the published GNSS/INS poses ranged from 356 to 825~m, with a median of 489~m. Its downward-looking camera recorded $1600\times1200$ images at 25~Hz. We use the 2,706 query images released at $800\times600$ pixels and 1~Hz. The dataset also supplies rendered reference views, depth, and reference poses with six degrees of freedom. We run odometry on the query images alone, using the offline-rectified grayscale query images and their corresponding pinhole calibration for feature extraction and geometric verification. We use the published roll, pitch, and yaw as ground truth for orientation evaluation.

Table~\ref{tab:rovo-config} lists the ROVO settings for the flight experiments. We set the keyframe support threshold close to or equal to the per-image feature budget so that almost every input frame becomes a keyframe. The number of matches surviving geometric verification usually falls below this threshold. In the seven flight runs, 99.99\% of input frames became keyframes. This schedule calls rotation averaging and inter-window alignment and fusion more frequently to optimize the orientation of each incoming frame. This is practical because these operations together took an average of only 8.6~ms per window update in MATLAB R2023b on an Intel Core i5-12600KF.

\begin{table*}[!tbp]
\centering\footnotesize
\captionstyle{center}
\setlength{\abovecaptionskip}{6pt}
\setlength{\belowcaptionskip}{3pt}
\setlength{\tabcolsep}{3pt}
\caption{ROVO run-specific settings for the flight experiments. Feature count is the per-image budget, descriptor threshold is the MATLAB matching threshold, and keyframe support is the threshold applied to the largest surviving match count.}
\label{tab:rovo-config}
\begin{tabular}{@{}cccccc@{}}
\toprule
Run & Image condition & Features/image & Descriptor threshold & Keyframe support & Window/stride \\
\midrule
Bell412-3 & Original & 1,000 & 1 & 800 & 10/1 \\
Bell412-3 & $3\times$ focal length & 1,000 & 1 & 1,000 & 10/1 \\
Bell412-4 & Original & 1,000 & 1 & 800 & 10/1 \\
Bell412-4 & $3\times$ focal length & 800 & 1 & 800 & 10/1 \\
Bell412-5 & Original & 1,000 & 2 & 800 & 10/1 \\
Bell412-5 & $3\times$ focal length & 1,000 & 5 & 1,000 & 10/1 \\
VPAIR & Offline rectified & 1,000 & 1 & 800 & 10/1 \\
\bottomrule
\end{tabular}

\vspace{2pt}
\begin{minipage}{0.96\textwidth}
\footnotesize
All runs use SIFT with contrast and edge thresholds of 0.005 and 20, respectively, and the joint rotation selector. Shared inter-window settings are $\tau_{\mathrm{align}}=4^\circ$, $\sigma_{\mathrm{align}}=3^\circ$, $\sigma_{\mathrm{fuse}}=4^\circ$, $\theta_{\max}=2^\circ$, and $(I_{\min},I_{\max})=(0.1,4)$.
\end{minipage}
\end{table*}

\subsection{Relative Rotation Error}

We evaluate relative rotation error (RRE) with the rotation-angle relative pose error convention of \texttt{evo}~\cite{grupp2017evo}. Estimated and reference poses are associated by nearest timestamp within 0.05~s. We report the root mean square of the relative rotation angles in degrees.

\begin{table}[!tbp]
\centering\footnotesize
\captionstyle{center}
\setlength{\abovecaptionskip}{6pt}
\setlength{\belowcaptionskip}{3pt}
\setlength{\tabcolsep}{3.5pt}
\caption{Bell412 RRE RMSE ($^\circ$) on shared pairs. Altitude is the maximum recorded for each flight. Bold indicates the lowest error.}
\label{tab:bell_rre}
\begin{tabular}{@{}ccccc@{}}
\toprule
Flight & Alt. (m) & ORB-SLAM3 & DPVO & ROVO \\
\midrule
\multicolumn{5}{c}{Original focal length}\\
Bell412-3 & 154 & 0.511 & \textbf{0.510} & 0.511 \\
Bell412-4 & 112 & 0.432 & \textbf{0.428} & 0.444 \\
Bell412-5 & 171 & 0.600 & \textbf{0.594} & 0.687 \\
\midrule
\multicolumn{5}{c}{Triple focal length}\\
Bell412-3 & 154 & 1.337 & 2.703 & \textbf{0.918} \\
Bell412-4 & 112 & 2.129 & 2.160 & \textbf{0.652} \\
Bell412-5 & 171 & 2.403 & 0.984 & \textbf{0.789} \\
\bottomrule
\end{tabular}
\end{table}

On Bell412 each imaging condition uses adjacent reference frames with distinct associated poses in all three systems. On VPAIR the primary per-method score uses adjacent released queries with both poses available. ROVO pairs must also lie in the same tracking segment since a restart introduces an independent orientation reference. We report coverage separately and repeat the VPAIR comparison on the intersection of the three sets of valid pairs.

\subsection{Results on MUN-FRL}\label{sec:scaled}
\begin{figure*}[!tb]
\centering
\includegraphics[width=\textwidth]{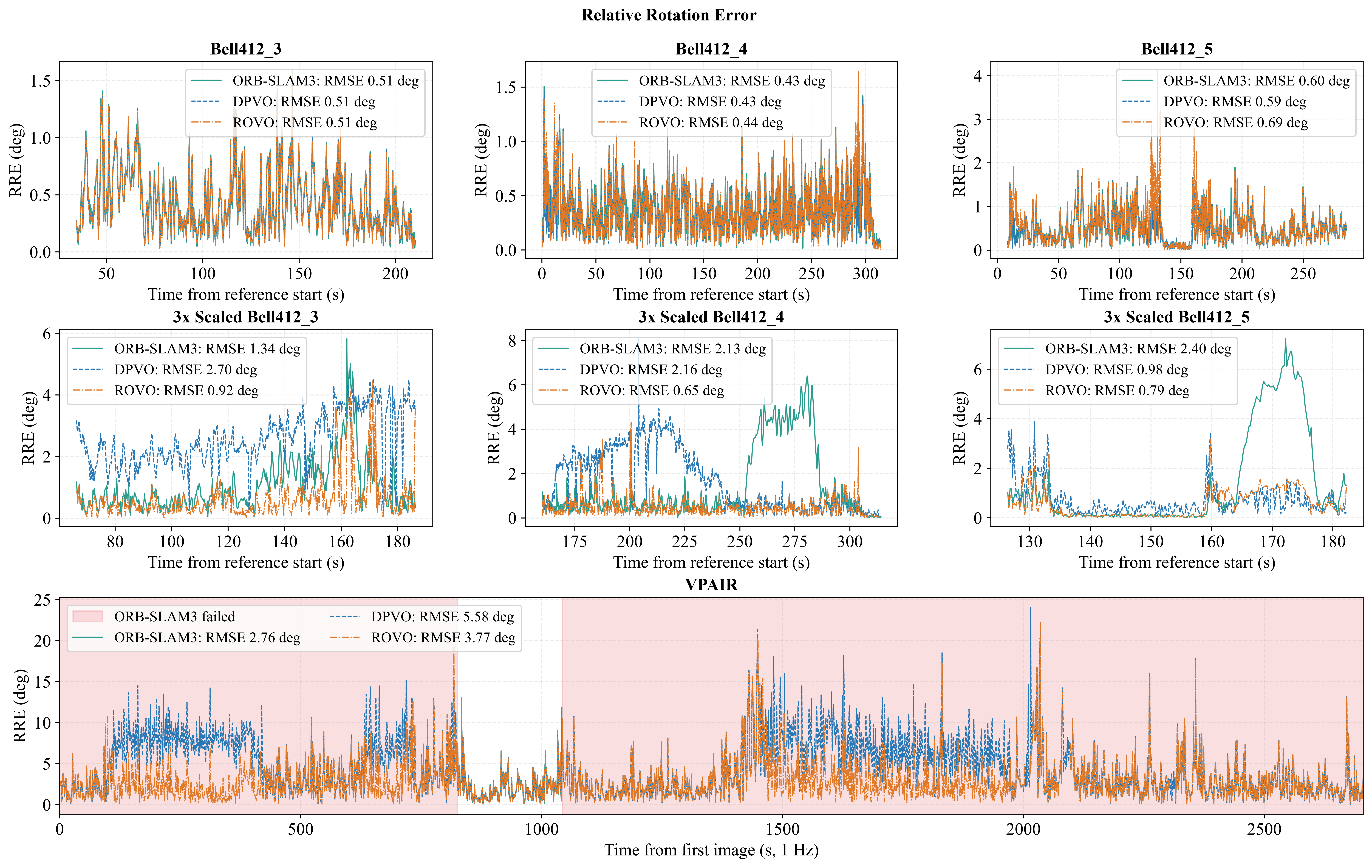}
\caption{Relative rotation errors of ORB-SLAM3, DPVO, and ROVO with joint rotation selection. Top and middle rows show shared Bell412 pairs at the original and triple focal lengths. The bottom row uses each method's valid adjacent VPAIR pairs. Shading indicates missing ORB-SLAM3 poses. Legend values are per-sequence RMSEs in degrees.}
\label{fig:fig8}
\end{figure*}

Table~\ref{tab:bell_rre} and Figure~\ref{fig:fig8} report the results for the original and reprojected Bell412 images.

On the original images, the three Bell412 error traces often rise and fall together, with DPVO giving the lowest sequence RMSE in each case. The broader view provides more opportunities to maintain spatially distributed tracks across frames. With these tracks, map or patch constraints can help stabilize the rotation estimate as part of joint pose estimation. The error traces alone cannot distinguish this effect from differences in the solvers and track selection.

On the shared frame pairs in the narrow-FoV configuration, ROVO reduces relative rotation RMSE by 31.3\% on Bell412-3, 69.4\% on Bell412-4, and 19.7\% on Bell412-5 compared with the better of ORB-SLAM3 and DPVO on each sequence. This advantage is consistent with the focal-length simulation (Figure~\ref{fig:fig6}), in which translation-direction error generally increases as the FoV narrows. In methods that jointly estimate pose and scene structure, errors in translation direction can propagate through estimated landmark or patch depths into subsequent pose estimates, including rotation. ROVO averages relative rotations across overlapping views and carries forward only orientations. By avoiding this path of error propagation, ROVO can maintain more accurate orientation estimates when translation direction is poorly constrained.

\subsection{Results on VPAIR}
On the higher-altitude VPAIR sequence, ROVO returned orientations for 2,681 of the 2,706 queries (99.1\%) across eight tracking segments (Table~\ref{tab:vpair_rre}). As the altitude sweep in Figure~\ref{fig:fig5} illustrates, greater scene depth weakens translational parallax at a fixed baseline and makes translation direction more sensitive to image noise. ROVO estimates and averages relative rotations from geometrically verified feature matches. It starts each tracking segment without initializing a 3D map or estimating a translation scale, and its orientation-only state avoids propagating translation errors through estimated depths. ORB-SLAM3 restarted 21 times during the run and returned 217 poses (8.0\%) in a single interval. Its monocular initialization requires enough parallax to triangulate an initial map~\cite{ORBSLAM3_TRO}. DPVO returned poses for all 2,706 queries using learned patch tracking and joint refinement of camera poses and patch depths~\cite{NEURIPS2023_7ac484b0}.

Across the VPAIR sequence, ROVO reduced relative rotation RMSE by 32.5\% compared with DPVO on their shared frame pairs. ROVO also achieved the lowest RMSE among the three methods on the short segment that ORB-SLAM3 successfully tracked.

\begin{table*}[!tbp]
\centering
\captionstyle{center}
\setlength{\abovecaptionskip}{6pt}
\setlength{\belowcaptionskip}{3pt}
\caption{VPAIR tracking coverage and RRE RMSE ($^\circ$) against the published attitude reference. The final column uses the same 216 adjacent query pairs for all methods. Bold marks the highest coverage and the lowest common-pair error.}
\label{tab:vpair_rre}
\begin{tabular}{cccccc}
\toprule
Method & Poses & Coverage (\%) & Valid pairs & Own-pair RMSE & Common-pair RMSE \\
\midrule
ORB-SLAM3$^{\dagger}$ & 217 & 8.0 & 216 & 2.761 & 2.761 \\
DPVO & \textbf{2,706} & \textbf{100.0} & 2,705 & 5.583 & 2.639 \\
ROVO & 2,681 & 99.1 & 2,673 & 3.767 & \textbf{2.631} \\
\bottomrule
\multicolumn{6}{c}{\footnotesize $^{\dagger}$ORB-SLAM3 covers 8.0\% of the flight. Its two RMSEs use the same interval.}\\

\end{tabular}
\end{table*}

\subsection{Ablation Study}\label{sec:ablation}
We examine the contributions of adaptive rotation selection and robust inter-window fusion. Table~\ref{tab:ablation} compares four ROVO variants on the same adjacent reference pairs within each flight condition. Pairs spanning a tracking-segment boundary are excluded. The five-point, eigensolver, and Hybrid Join variants use the proposed fusion, while Overwrite uses Hybrid Join with direct replacement of the aligned window estimates.

\begin{table*}[!tbp]
\centering\footnotesize
\captionstyle{center}
\setlength{\abovecaptionskip}{6pt}
\setlength{\belowcaptionskip}{3pt}
\setlength{\tabcolsep}{7pt}
\caption{Ablation results in RRE RMSE ($^\circ$). All four variants use the same adjacent reference pairs in each row. Bold indicates the lowest error.}
\label{tab:ablation}
\begin{tabular}{@{}ccccc@{}}
\toprule
Sequence & Five-point & Eigensolver & Hybrid Join & Overwrite \\
\midrule
\multicolumn{5}{c}{Original focal length}\\
Bell412-3 & 0.5114 & 0.6107 & \textbf{0.5109} & 0.5137 \\
Bell412-4 & 0.4445 & 1.0842 & \textbf{0.4444} & 0.4536 \\
Bell412-5 & \textbf{0.6831} & 0.8924 & 0.6870 & 0.7044 \\
\midrule
\multicolumn{5}{c}{Triple focal length}\\
Bell412-3 & \textbf{0.8735} & 1.0592 & 0.9183 & 1.0681 \\
Bell412-4 & \textbf{0.6321} & 0.9018 & 0.6525 & 0.8033 \\
Bell412-5 & 0.8040 & 0.8276 & \textbf{0.7796} & 0.8230 \\
\midrule
\multicolumn{5}{c}{Higher-altitude flight}\\
VPAIR$^{\dagger}$ & 3.7925 & 3.9232 & \textbf{3.7665} & 3.7884 \\
\bottomrule
\end{tabular}

\vspace{2pt}
\begin{minipage}{0.96\textwidth}
\footnotesize
The narrow-FoV Bell412-5 Hybrid Join result uses the completed rerun. $^{\dagger}$The fixed-solver VPAIR runs extract features before rectification, while both Hybrid Join variants use offline-rectified images.
\end{minipage}
\end{table*}

\subsubsection{Adaptive Rotation Selection}
Hybrid Join consistently yields lower error than the fixed eigensolver and remains close to the five-point baseline. Its advantage over five-point varies across sequences, with five-point retaining the lowest error on some Bell412 flights. This pattern supports retaining the five-point candidate while admitting an eigensolver refinement when the observed views support it. The refinement is sensitive to the distribution of viewing directions and to perturbations in the matched bearings, so its contribution depends on the image pair and the surrounding constraints.

The candidate records show that Hybrid Join selects eigensolver refinements more frequently under the narrower FoV. When both candidates pass verification, the selected refinement usually has the lower reference rotation error. These observations indicate that the selector responds to changes in viewing geometry through geometric support and multi-view agreement. The altitude and noise sweeps in Section~\ref{sec:sim} explain the motivation for this adaptation, since both factors affect the sensitivity of the underlying solvers. The effect on trajectory accuracy also depends on how the selected measurements interact with the remaining rotation graph, which helps explain the varying gains over five-point.

\subsubsection{Robust Inter-Window Fusion}
With Hybrid Join as the front end, robust fusion gives lower error than direct overwrite in every flight condition, with the largest benefit under the narrower FoV. Both variants use the same window schedule and overlap-based alignment. As the window advances, changes in its image constraints can produce different estimates of an orientation already present in the trajectory. Direct replacement transfers this variation into the stored poses. The larger improvement under the narrower FoV is consistent with the increased sensitivity of rotation estimates when the viewing directions are less widely distributed.

The proposed update combines the aligned window estimate with the stored orientation using their information weights. Observation quality and rotational residuals control the contribution of each update, while the correction bound limits the effect of an inconsistent window estimate. Earlier window information therefore continues to influence the trajectory as new constraints arrive. Bounding the accumulated information also preserves the influence of later observations. Since the stored orientations guide subsequent alignment and candidate selection, retaining a robust history can benefit later window estimates as well as the current update.

\section{Conclusion}\label{sec:conclusion}
ROVO combines adaptive rotation selection, sliding-window rotation averaging, and robust inter-window fusion to estimate camera orientations from monocular aerial images. Simulations illustrate the benefit of estimating rotation when translation direction is poorly constrained. Flight experiments show accuracy comparable to ORB-SLAM3 and DPVO under the original Bell412 settings, with improved accuracy under a narrower FoV and on higher-altitude VPAIR. The ablations support adapting rotation measurements through multi-view agreement and retaining information across windows. Future work will integrate ROVO with inertial and translation measurements for full-pose navigation.

%%%%%%%%%%%%%%%%%%%%%%%%%%%%%%%%%%%%%%%%%%%%%%%%%%%%%%%
%%% Acknowledgements. ��л
%%%%%%%%%%%%%%%%%%%%%%%%%%%%%%%%%%%%%%%%%%%%%%%%%%%%%%%

\Acknowledgements{This work was supported in part by the National Natural Science Foundation of China under Grant 12472189, the Science and Technology Innovation Program of Hunan Province under Grant 2022RC1196, and the Fund of the National Key Laboratory of Advanced Propulsion Technology under Grant No.~25-GC-ZZKY-021.}

%%%%%%%%%%%%%%%%%%%%%%%%%%%%%%%%%%%%%%%%%%%%%%%%%%%%%%%
%%% Reference section. �ο�����
%%% Citation in the content using "some words~\cite{1,2}".
%%% ~ is needed to make the reference number is on the same line with the word before it.
%%% Bibliography entries and formatting are supplied by refs.bib and scts.bst.
\bibliographystyle{scts}
\bibliography{refs}

%%%%%%%%%%%%%%%%%%%%%%%%%%%%%%%%%%%%%%%%%%%%%%%%%%%%%%%

%%%%%%%%%%%%%%%%%%%%%%%%%%%%%%%%%%%%%%%%%%%%%%%%%%%%%%%
%%% Appendix sections. ��¼�½�, �Ǳ�ѡ
%%%%%%%%%%%%%%%%%%%%%%%%%%%%%%%%%%%%%%%%%%%%%%%%%%%%%%%
%\begin{appendix}
%\section{Name}

%\end{appendix}

\end{document}